\documentclass{article}

\usepackage{microtype}
\usepackage{color}
\usepackage{graphicx}
\usepackage{subcaption}
\usepackage{booktabs} 
\usepackage[margin=1in]{geometry}

\usepackage{hyperref}

\newcommand{\red}[1]{\textcolor{black}{#1}}

\usepackage{tikz}
\usetikzlibrary{positioning, arrows.meta, fit, backgrounds, calc}

\usepackage{amsmath}
\usepackage{amsfonts}
\usepackage{amssymb}
\usepackage{amsthm}
\usepackage{float}
\usepackage{url}
\usepackage[toc,page]{appendix}
\usepackage{tabularx}
\usepackage{bbm}
\usepackage{multirow}

\theoremstyle{plain}

\theoremstyle{definition}

\theoremstyle{remark}

\title{Latent Chain-of-Thought Improves Structured-Data Transformers}

\author{
Carson Dudley \\
\texttt{cdud@umich.edu}
\and
Samet Oymak \\
\texttt{oymak@umich.edu}
}

\date{\today}

\begin{document}

\maketitle







\begin{abstract}
Chain-of-thought and more broadly test-time compute are known to augment the expressive capabilities of language models and have led to major innovations in reasoning. Motivated by this success, this paper explores latent chain-of-thought as well as the impact of depth and looping for time-series and tabular data. We propose a recurrent scheme in which a structured-data transformer, after an initial forward pass, compresses its query-position hidden states into feedback tokens that are appended to the input and processed again, allowing multiple rounds of latent computation before prediction. We compare CoT models against a same-depth no-CoT baseline, a deeper baseline matched to the CoT model in effective depth, and a looped transformer with weight-tied recurrence but no additional chain-of-thought tokens. Across 36 datasets in time-series forecasting and tabular prediction, latent chain-of-thought improves over the baseline on 7/9 time-series datasets (+12.63\% average gain) and 23/27 tabular datasets (+3.25\% average gain), with CoT models performing best on average in both settings. We also show that the benefit of CoT extends to pretrained foundation models: applying latent CoT to nanoTabPFN, a small open-source tabular foundation model, improves its performance above the much larger TabPFN-v2 on TabArena. Together, these results demonstrate that chain-of-thought is a useful axis for scaling test-time compute for structured data.
\end{abstract}

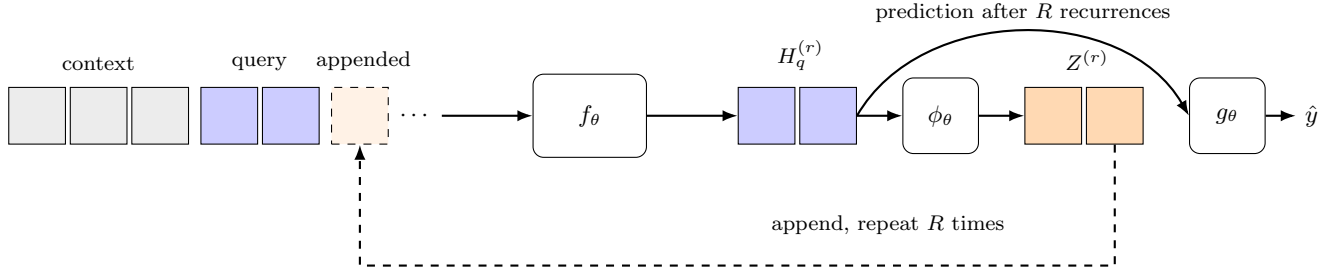
\begin{figure*}[t]
\centering
\hspace*{-0.3cm}
\begin{tikzpicture}[
  every node/.style={font=\small},
  ctx/.style={draw, rectangle, minimum width=0.75cm, minimum height=0.75cm, fill=gray!15},
  qry/.style={draw, rectangle, minimum width=0.75cm, minimum height=0.75cm, fill=blue!20},
  fb/.style={draw, rectangle, minimum width=0.75cm, minimum height=0.75cm, fill=orange!30},
  slot/.style={draw, rectangle, minimum width=0.75cm, minimum height=0.75cm, dashed, fill=orange!10},
  block/.style={draw, rectangle, rounded corners, minimum width=1.5cm, minimum height=1.1cm, fill=white},
  arr/.style={-{Latex[length=2mm]}, thick},
]

\node[ctx] (c1) at (0,0) {};
\node[ctx, right=0.05cm of c1] (c2) {};
\node[ctx, right=0.05cm of c2] (c3) {};
\node[qry, right=0.15cm of c3] (q1) {};
\node[qry, right=0.05cm of q1] (q2) {};

\node[slot, right=0.15cm of q2] (s1) {};
\node[right=0.02cm of s1] (dots) {$\cdots$};
\node[above=0.1cm of c2] {\footnotesize context};
\node[above=0.1cm of q1, xshift=0.4cm] {\footnotesize query};
\node[above=0.1cm of s1, xshift=0.05cm] {\footnotesize appended};

\node[block, right=1.2cm of dots] (tf) {$f_\theta$};

\node[qry, right=1.2cm of tf] (h1) {};
\node[qry, right=0.05cm of h1] (h2) {};
\node[above=0.1cm of h1, xshift=0.45cm] {\footnotesize $H^{(r)}_q$};

\node[block, right=0.6cm of h2, minimum width=1.0cm, minimum height=1.0cm] (mlp) {$\phi_\theta$};

\node[fb, right=0.6cm of mlp] (z1) {};
\node[fb, right=0.05cm of z1] (z2) {};
\node[above=0.1cm of z1, xshift=0.45cm] {\footnotesize $Z^{(r)}$};

\draw[arr] (dots.east) -- (tf.west);
\draw[arr] (tf) -- (h1);
\draw[arr] (h2) -- (mlp);
\draw[arr] (mlp) -- (z1);

\draw[arr, dashed] (z2.south) -- ++(0,-1.6)
  node[below, midway, xshift=-3cm] {\footnotesize append, repeat $R$ times}
  -| (s1.south);

\node[block, right=0.6cm of z2, minimum width=1.0cm, minimum height=1.0cm] (g) {$g_\theta$};
\node[right=0.4cm of g] (yhat) {$\hat{y}$};
\draw[arr] (h2.east) to[bend left=60] node[above, midway] {\footnotesize prediction after $R$ recurrences} (g.west);
\draw[arr] (g) -- (yhat);
\end{tikzpicture}
\caption{\textbf{\red{Latent} chain-of-thought for structured data.} The transformer $f_\theta$ runs on a sequence of context tokens, query tokens, and (after the first pass) appended feedback tokens. Query-position hidden states $H^{(r)}_q$ are compressed by an MLP $\phi_\theta$ into feedback tokens $Z^{(r)}$, which are appended to the sequence for the next pass. After $R$ recurrences, the \red{prediction} head $g_\theta$ maps from the hidden states to the prediction $\hat{y}$.}
\label{fig:method}
\end{figure*}

\section{Introduction}

One of the main vehicles for recent progress in large language models has been scaling test-time compute. Reasoning models such as OpenAI's o1 \cite{openai2024o1}, DeepSeek-R1 \cite{deepseek2025r1}, and their successors demonstrate that allocating more computation per query can substitute for, and sometimes outperform, scaling parameters \cite{snell2024scaling}. One of the main mechanisms for scaling test-time compute is chain-of-thought reasoning: inducing the model to generate intermediate reasoning tokens in natural language to think through the problem \cite{wei2023chain}.

It is not clear whether this reasoning has to happen in natural language. Most tokens in a chain of thought serve textual coherence rather than computation, and decoding to a discrete token at each step discards most of the information in the model's hidden state. \cite{hao2025coconut} address this directly with Coconut, which feeds the model's last hidden state back as the next input embedding rather than decoding it to a single word, replacing discrete token-based chain-of-thought with a chain of latent thoughts. \cite{geiping2025scaling} achieve a similar goal through recurrent depth: rather than extending the sequence, the model iteratively applies a shared transformer block to refine its internal latent state before producing an output. This enables additional test-time computation without requiring chain-of-thought supervision or longer context. \red{\cite{gozeten2025continuous} further show that latent chain-of-thought enables parallel exploration of multiple reasoning paths, demonstrating additional flexibility of this paradigm for test-time scaling.} Both of these approaches build on earlier work on weight-tied depth recurrence in the Universal Transformer \cite{dehghani2019universal} and looped transformers \cite{giannou2023looped}, as well as papers showing that even content-free tokens can be used to allocate additional forward-pass computation \cite{goyal2024think}.

All of this work has focused on language. Yet the same learning setup (a transformer doing in-context learning) is now used by state-of-the-art models for structured data. In tabular prediction, TabPFN \cite{hollmann2023tabpfn} introduced in-context learning (ICL) over labeled rows, and its successors TabPFNv2 \cite{tabpfnv2} and TabICL \cite{qu2025tabicl} use alternating row and column attention to scale this paradigm to medium and large tables, challenging the long-standing dominance of gradient-boosted trees \cite{xgboost}. In time-series forecasting, transformer-based foundation models such as Chronos \cite{ansari2024chronos}, TimesFM \cite{timesfm}, and Moirai \cite{moirai}, as well as some domain-specific foundation models \cite{fincast, mantis}, have become the standard. In both tabular prediction and time-series forecasting the model is a transformer that attends over a structured set of observations and produces predictions in a single forward pass---the same setting where adding latent computation has paid off in \red{language modeling}. Recent mechanistic analysis of tabular foundation models suggests that iterative computation may already be doing significant work inside these architectures. \cite{balef2026layerenoughunderstandinginference} probes state-of-the-art tabular foundation models (TabPFN, TabICL, LimiX) and find substantial depth-wise redundancy, with middle and late layers performing overlapping refinement of the residual stream rather than learning distinct transformations. They show that a single transformer block looped six times recovers the performance of a full six-layer model with only 20\% of the parameters. This suggests that depth in tabular foundation models is primarily a vehicle for iterative refinement---which raises the question of whether explicit test-time iteration can extract further gains.

In this paper, we ask whether latent chain of thought can transfer to structured data. Our approach gives a structured-data transformer multiple rounds of latent computation before it makes a prediction. After an initial forward pass, we take the hidden states at the prediction positions, compress them through a small MLP into feedback tokens, append those tokens to the model's input representation, and run the transformer again. To enable controlled comparisons, we train each model from scratch on a single dataset rather than pretraining a separate foundation model for each recurrence variant. We compare against (i) a no-recurrence baseline at the same depth, (ii) a deeper baseline matched in total compute, isolating the contribution of latent recurrence from the confounds of pretraining scale and parameter count, and (iii) looped-transformer baselines that share weights across passes but lack the explicit feedback-token mechanism \red{(i.e., recurrent passes with no CoT tokens)} \cite{giannou2023looped}.

Training from scratch makes the comparison between mechanisms clean, but does not demonstrate whether latent CoT still helps once a foundation model has been pretrained at scale. To address this, we additionally pretrain nanoTabPFN \cite{pfefferle2025nanotabpfn}, a small open-source implementation of the TabPFN-v2 architecture, with latent CoT. On the TabArena small-table benchmark used in the nanoTabPFN paper \cite{tabarena}, our pretrained nanoTabPFN-CoT models surpass full TabPFN-v2 despite using only a 3-layer transformer with a small embedding dimension, suggesting that latent CoT can partially compensate for model scale even in the pretrained regime.

We adapt latent chain of thought to tabular prediction and time-series forecasting, evaluate it on nine time-series datasets from LTSF-9 \cite{zeng2022lstf} and a 27-dataset subset of OpenML that was used in TabPFN evaluations \cite{openml}, and characterize when our approach improves prediction relative to both same-depth and matched-compute baselines. We additionally evaluate latent CoT in the pretrained foundation-model setting on TabArena \cite{tabarena}.

\begin{table*}[h]
\centering
\caption{
Aggregate benchmark performance across structured-data tasks.
``Best CoT'' uses the CoT depth selected using validation performance from
$\{1,2,4\}$ independently for each dataset, ``Best Looped'' uses the same validation approach.
Average rank is computed per dataset across baseline, deeper, best CoT, and best looped.
Average gain is reported relative to the same-depth baseline, with positive values indicating improvement.
}
\label{tab:aggregate_results}
\resizebox{0.75\linewidth}{!}{
\begin{tabular}{llccc}
\toprule
Domain & Method & Avg. Rank $\downarrow$ & Wins vs. Baseline & Avg. Gain vs. Baseline $\uparrow$ \\
\midrule

\multirow{3}{*}{Time series}
& Baseline & 3.00 & -- & -- \\
& Deeper & 2.94 & 4 / 9  & $-5.45\% \pm 4.82\%$ \\
& Best CoT & 1.56 & 7 / 9 & $+12.63\% \pm 5.08\%$ \\
& Best Looped & 2.50 & 7 / 9 & $+2.77\% \pm 1.05\%$ \\

\midrule

\multirow{3}{*}{Tabular}
& Baseline & 2.89 & -- & -- \\
& Deeper & 2.78 & 14 / 27 & $+0.66\% \pm 0.63\%$ \\
& Best CoT & 1.59 & 23 / 27 & $+3.25\% \pm 0.88\%$ \\
& Best Looped & 2.74 & 14 / 27 & $-0.39\% \pm 0.80\%$ \\

\bottomrule
\end{tabular}
}
\label{result_table}
\end{table*}

\section{Methods}

\subsection{Latent chain-of-thought for structured data transformers}

We study whether structured-data transformers can benefit from latent chain-of-thought. Let $x$ denote the structured input, and let $q$ denote the prediction positions: query rows in tabular prediction or future-patch query tokens in time-series forecasting. A standard transformer computes hidden states $H^{(0)} = f_\theta(x),$ and predicts from the hidden states at the query positions using a prediction head, $\hat{y}^{(0)} = g_\theta(H^{(0)}_q).$ Our method adds recurrent passes through the same transformer. At recurrence step $r$, we extract the query-position hidden states $H^{(r)}_q$, map them through a two-layer MLP $Z^{(r)} = \phi_\theta(H^{(r)}_q)$, and append the resulting tokens to the embedded input sequence: $E^{(r+1)} = [E^{(0)}; Z^{(0)}; \ldots; Z^{(r)}]$, where $E^{(0)}$ denotes the embedding of $x$. The transformer is then run again on the sequence $H^{(r+1)} = f_\theta\big(x, Z^{(0)}, \ldots, Z^{(r)}\big).$ After $R$ recurrences, the final prediction is decoded from the original query positions, $\hat{y} = g_\theta(H^{(R)}_q).$

For tabular tasks, we use the three-stage architecture of TabICL \cite{qu2025tabicl}. Each scalar cell is embedded as a token, with learned embeddings added to distinguish columns. The input consists of context rows $(X_c, y_c)$ and query rows $X_q$. Context labels are in a label column, while query rows receive a learned no-label token. A column-wise transformer $\mathrm{TF}_{\mathrm{col}}$ first attends across samples within each feature, a row-wise transformer $\mathrm{TF}_{\mathrm{row}}$ then attends across features within each row to produce a single embedding per row, and an in-context learning transformer $\mathrm{TF}_{\mathrm{icl}}$ attends across rows to predict query labels from labeled context rows. Recurrence is applied at the $\mathrm{TF}_{\mathrm{icl}}$ stage: the query positions $q$ are the query-row positions in $\mathrm{TF}_{\mathrm{icl}}$, and feedback tokens are appended to the row-embedding sequence before $\mathrm{TF}_{\mathrm{icl}}$ is rerun. The first two stages are run once per query and their outputs are reused across recurrences. Predictions are trained with cross-entropy for classification and \red{root} mean squared error for regression.

For time-series tasks, we use a patch-based forecasting transformer \cite{nie2023patchtst}. A (potentially multivariate) context window $X_{1:T} \in \mathbb{R}^{T \times C}$ is divided into patches and embedded into a sequence of context tokens. We append learned query tokens corresponding to future patches and run a standard transformer over the combined sequence. The hidden states at the future-query positions are decoded into quantile forecasts for the prediction horizon. During chain-of-thought, the future-query hidden states are appended to the input before rerunning the same transformer stack. The final forecast is decoded after $R$ recurrences. We train forecasts using the quantile loss over quantile levels $\{0.1,0.3,0.5,0.7,0.9\}$.

We train each model from scratch on each dataset with a fixed chain-of-thought length $R_{\mathrm{train}} \in \{0,1,2,4\}$ and evaluate at multiple recurrence depths $R \in \{0,1,2,4,8\}$. For each task, we compare against three families of baselines, each designed to isolate a specific aspect of our experiment. The first baseline is a same-depth $L$-layer transformer without CoT ($R=0$). This controls for parameter count and isolates the effect of additional test-time computation. Next, we train a deeper baseline, a $2L$-layer transformer trained without recurrence, matching a CoT model at $R=1$ in effective forward-pass depth (and roughly in FLOPs) while doubling the parameter count. Comparing our CoT models to this baseline distinguishes gains from CoT from gains attributable simply to depth or capacity. Finally, we compare to looped baselines that share weights across passes but do not append additional tokens to the input. For these, a stack of $K$ transformer blocks is applied $M$ times for an effective depth of $K \cdot M$. We consider four configurations covering two regimes: weight-tied single-block looping in the style of \cite{giannou2023looped} ($K=1$, $M \in \{2,4\}$) and Universal-Transformer-style looping over a multi-block stack \cite{dehghani2019universal} ($K=4$, $M \in \{2,4\}$). Comparing CoT against the looped baselines isolates the contribution of the additional CoT tokens specifically, since both methods reuse the same weights across passes. All baselines are trained from scratch with the same optimizer, learning-rate schedule, and epoch budget as the CoT models.

\subsection{Extension to pretrained models}
\label{sec:pretrained}

The training from scratch protocol we describe for our main experiments allows for controlled comparisons between different configurations. However, in practice, the main application of structured data transformers is pretrained foundation models (e.g., \cite{hollmann2023tabpfn, tabpfnv2, ansari2024chronos} and many more). A concern is that the benefits of latent CoT may not transfer to the pretrained regime, where the model has already learned powerful in-context learning behavior from large-scale synthetic pretraining. To test this, we apply latent CoT to nanoTabPFN \cite{pfefferle2025nanotabpfn}, a small open-source implementation of the TabPFN-v2 architecture consisting of only a 3-layer transformer with embedding dimension 96. Although nanoTabPFN does not perform as well as TabPFN-v2, it performs surprisingly well on small tabular benchmarks relative to traditional machine learning baselines (e.g., tree-based methods \cite{xgboost}). nanoTabPFN is a useful testbed for studying whether latent CoT can compensate for model scale in pretrained tabular foundation models.

Adapting latent CoT to the TabPFN-v2 architecture requires some modification. Unlike the standard transformer setting in which one CoT step appends one token to the sequence, for a table with $d$ features and one label column, TabPFN-v2's architecture represents each row as $d+1$ vectors (one per feature plus one for the label) and alternates attention across features and across rows. A single CoT vector therefore does not fit into the row dimension; we have to append $d+1$ vectors per CoT step. In our configuration, the compressed query-position hidden state (i.e., the CoT token) occupies the label-column slot, and the $d$ feature column slots are filled with a learned placeholder embedding shared across CoT steps. Additional implementation details are in the appendix.

We evaluate using the protocol from the nanoTabPFN paper, which evaluates on a subset of TabArena containing small datasets with at most 10 features \cite{tabarena}. We pretrain nanoTabPFN with $R_\mathrm{train} \in \{1, 2, 4\}$, and compare performance to the original nanoTabPFN and full TabPFN-v2 results they report in \cite{pfefferle2025nanotabpfn}.

\section{Results}

\subsection{Latent chain-of-thought improves prediction for structured data transformers that are trained from scratch}

\begin{figure*}[t]
\centering
\includegraphics[width=0.8\linewidth]{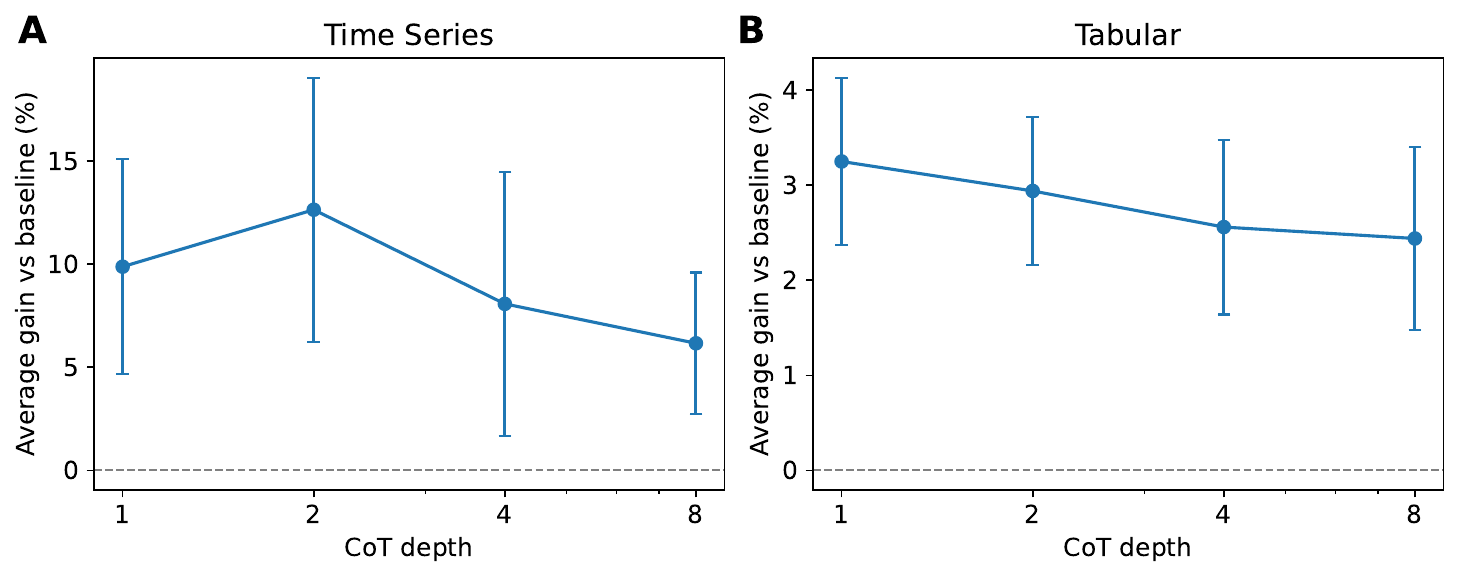}
\caption{
\textbf{Performance gains from latent chain-of-thought as a function of recurrence depth.} Each point is the mean across datasets of the per-dataset gain over the same-depth no-recurrence baseline at a fixed training/evaluation depth $R$ (in contrast to Table~\ref{tab:aggregate_results}, which selects $R$ per dataset on a validation split). \textbf{(A)} Time series: average gain in quantile loss (\%) across the nine LTSF datasets. The depth-8 point is obtained by training at $R_{\mathrm{train}}=4$ and increasing recurrences at evaluation only. \textbf{(B)} Tabular: average gain in classification accuracy (percentage points) across classification datasets in the OpenML subset. Dashed line indicates the same-depth baseline, error bars represent standard error.
}
\label{fig:cot_scaling}
\end{figure*}

Table \ref{tab:aggregate_results} reports aggregate performance across both domains. Selecting chain-of-thought depth per dataset based on average validation performance (i.e., we select the best CoT depth for the entire domain based on the average best validation performance), latent chain-of-thought improves over the same-depth baseline on 8 out of 9 time-series datasets ($+12.63\% \pm 5.08\%$ average gain in mean squared error) and 23 out of 27 tabular datasets ($+3.25\% \pm 0.88\%$ average gain). The same chain-of-thought models also outperform the depth-matched deeper baseline on average by 19.12\% on time series and 2.57\% on tabular, indicating that the gains are not simply a consequence of additional compute in the forward pass. The deeper baseline itself gives mixed results: it has strictly more capacity than the same-depth baseline, and on tabular tasks it yields a small average improvement ($+0.66\%$, winning 14/27 datasets), but on time series it actually underperforms the baseline on average ($-5.45\%$, winning only 4/9 datasets). Several of the benchmark datasets are small, and simply stacking more layers appears to sometimes encourage overfitting rather than useful additional computation. The CoT and looped models, by contrast, achieve more effective depth without adding parameters, and the resulting weight reuse acts as a form of regularization. Finally, the best CoT model also outperforms the best looped model by 7.74\% on time series forecasting and 3.82\% on tabular prediction, indicating that \red{both recurrence and explicit chain-of-thought tokens are necessary for maximum benefit}. Per-dataset results are in the appendix.

\paragraph{Scaling with chain-of-thought depth.}
Figure \ref{fig:cot_scaling} shows how performance varies with chain-of-thought depth. On time series, all tested depths $R \in \{1,2,4,8\}$ improve over the baseline, with average gains in quantile loss ranging from $+6.16\%$ at $R=8$ to a peak of $+12.63\%$ at $R=2$, followed by a gradual decline at $R=4$ ($+8.07\%$) and $R=8$ ($+6.16\%$). The depth-8 setting is obtained by training at $R_{\mathrm{train}}=4$ and increasing the number of recurrences only at evaluation, suggesting that the learned recurrent update generalizes beyond its training horizon. On tabular classification tasks, the trend is broadly similar but flatter: gains peak at $R=1$ ($+3.25$ pp) and decline gradually with depth ($+2.94$ at $R=2$, $+2.56$ at $R=4$, $+2.44$ at $R=8$), remaining positive throughout. The shape of both curves---a peak at modest depth followed by gradual decline---is broadly consistent with observations in language modeling that more chain-of-thought often helps but can plateau or degrade past a point \cite{cot_collapse}.

\paragraph{Latent recurrence vs. added depth and looping.} Comparing chain-of-thought to the deeper and looped baselines isolates the contribution of CoT tokens from raw compute and weight-tied recurrence. The deeper baseline matches a CoT model at $R=1$ in forward-pass depth but has roughly twice the parameter count. On tabular it gives a small average gain ($+0.66\%$) and on time series it actually hurts ($-5.45\%$), consistent with the small size of several benchmark datasets: the extra parameters provide capacity for overfitting without a corresponding mechanism for useful additional computation. The looped baselines, by contrast, share weights across passes like our CoT models but reapply the transformer to the same token sequence rather than appending intermediate representations. They improve modestly over the baseline on time series but underperform CoT on average (2.50 vs.\ 1.56 average rank on time series, 2.74 vs.\ 1.59 on tabular). Together, these comparisons suggest that the gains come from the interaction of two components: weight tying lets the model spend more compute while reusing parameters across passes, which acts as a form of regularization and avoids the overfitting that hurts the deeper baseline on small datasets, and the appended CoT tokens give the extra compute a place to store intermediate results, turning shared-weight recurrence from undifferentiated reapplication into a structured iterative computation.

\section{Latent CoT improves pretrained tabular foundation models}
\begin{figure}[t]
\centering
\includegraphics[width=0.6\linewidth]{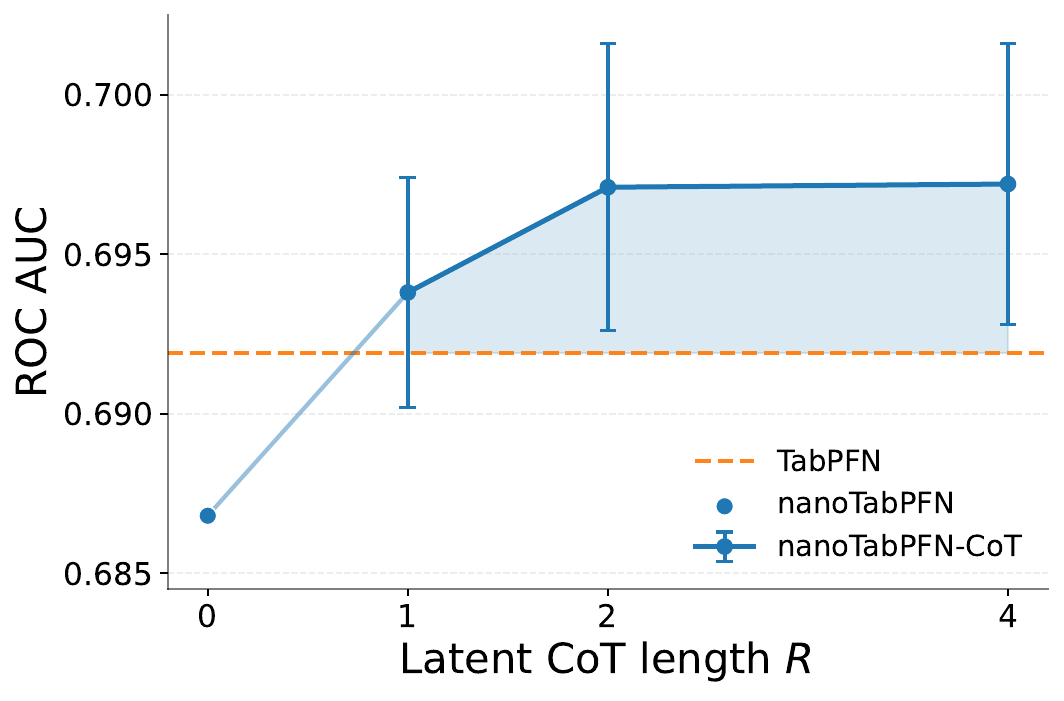}
\caption{
\textbf{Latent CoT improves pretrained nanoTabPFN and surpasses full TabPFN-v2 in the nanoTabPFN evaluation setting.}
ROC-AUC on the TabArena binary-classification benchmark used by \cite{pfefferle2025nanotabpfn}. The point at $R=0$ is nanoTabPFN without latent CoT from the original paper. Blue points show nanoTabPFN-CoT models pretrained with latent CoT length $R \in \{1,2,4\}$ across seeds, with error bars denoting standard errors. The dashed orange line marks the reported performance of full TabPFN-v2 under the same evaluation protocol. Shading indicates the region where nanoTabPFN-CoT exceeds TabPFN-v2.
}
\label{fig:nanotabpfn_cot}
\end{figure}

We have established that latent CoT often helps when training structured data transformers from scratch on single datasets. Our second evaluation focuses on whether latent CoT can improve the performance of pretrained transformers, where there may be less headroom for improvement. We pretrain nanoTabPFN with latent CoT using the same architecture, synthetic prior, and evaluation protocol as the original paper. Results are shown in figure \ref{fig:nanotabpfn_cot}. Latent CoT improves nanoTabPFN's AUC from $0.687$ (their original model) to $0.697$ at $R=2$ and $R=4$. The more relevant comparison is against larger pretrained tabular foundation models rather than nanoTabPFN at $R=0$. All nanoTabPFN-CoT variants achieve mean AUC above the reported performance of full TabPFN-v2 ($0.692$) on this benchmark, despite using a much smaller 3-layer model and a lightweight pretraining setup. For $R \geq 2$, the improvement is clearly separated from the TabPFN-v2 baseline, while the $R=1$ setting overlaps within standard error. Overall, these results suggest that latent CoT can partially compensate for model scale in pretrained tabular transformers.

The absolute improvement (roughly one ROC-AUC point) is modest in magnitude compared to the from-scratch results, but this is common for modern tabular foundation models. On benchmarks such as TabArena, absolute differences in performance are often small, so evaluation often emphasizes rank- and ELO-based comparisons rather than absolute scores (e.g., \cite{tabpfnv3}). In this setting, a one-point gain arising from a mechanism change is a meaningful improvement.

\section{Discussion}
Our results show that latent chain-of-thought can transfer from language modeling to structured-data transformers. Across nine LTSF time-series datasets and 27 OpenML tabular datasets, giving a transformer multiple rounds of latent computation before it makes a prediction outperforms a same-depth no-CoT baseline, a depth-matched deeper baseline, and weight-tied looped transformers on the majority of datasets. The fact that CoT beats all three baselines demonstrates that the gains are not attributable only to compute, parameter count, or weight tying, but to the mechanism of compressing query-position hidden states into feedback tokens and reprocessing them. The looped baselines are particularly informative since they share weights across passes but reapply the transformer to the same token sequence rather than appending an intermediate representation. Their underperformance suggests that the benefit of CoT is not iterative refinement of the hidden state per se, but the ability to write intermediate computation into a sequence that subsequent passes can attend to.

The shapes of the depth-scaling curves are also informative. On both tabular and time-series tasks, performance improves over the baseline at every tested depth but peaks at a relatively shallow recurrence count and declines gradually thereafter. Time series peaks at $R=2$ ($+12.63\%$) and tabular peaks at $R=1$ ($+3.25$ pp), with both curves remaining above the baseline through $R=8$. The $R=8$ point is obtained by training at $R_{\mathrm{train}}=4$ and increasing CoT depth only at evaluation, which indicates that the CoT update can generalize beyond its training horizon. The gradual decline at greater depths is similar to language modeling where more chain-of-thought helps up to a certain point, but then can plateau or degrade performance.

The from-scratch results raise the natural question of whether these gains survive into the pretrained regime, where structured data foundation models live. Our nanoTabPFN experiments suggest they do: pretraining a 3-layer TabPFN-v2-style model with latent CoT lifts performance above full TabPFN-v2 on the TabArena small-table benchmark, despite a much smaller parameter count and a lightweight pretraining setup. The effect size is smaller than in the from-scratch setting, which is unsurprising---headroom shrinks once a model has been pretrained at scale on tasks closely matched to the evaluation distribution---but the direction is consistent, and the improvement is meaningful in a benchmark where rank- and ELO-based comparisons are often used because absolute gaps between modern tabular foundation models are typically small \cite{tabpfnv3}. The fact that this relatively minimal adaptation already produces a measurable gain suggests that latent CoT can still provide benefit to pretrained foundation models.

Several limitations remain. The chain-of-thought depth $R$ is fixed per model rather than chosen adaptively per input, even though the non-monotonic performance curve suggests that the optimal depth varies across instances. An adaptive-CoT mechanism similar to PonderNet \cite{banino2021pondernet} would be a useful next step. We also do not characterize how CoT interacts with distribution shift, which is an important direction given that structured-data foundation models are typically deployed on data that differs from their pretraining distribution. Our pretrained-model evaluation is restricted to a single small architecture (nanoTabPFN) on a single benchmark (TabArena small tables). Whether the gains hold for larger pretrained models such as TabPFN-v2 itself, or for time-series foundation models, remains open. Finally, it would be insightful to mechanistically characterize what the feedback tokens actually encode---whether they represent residual error, prediction uncertainty, or something else.

\bibliographystyle{unsrt}

\appendix

\section{Architecture and training details}
\label{app:training}

All models are trained from scratch on each dataset using the same optimizer and schedule. We use AdamW with learning rate $3 \times 10^{-4}$, cosine annealing, weight decay $10^{-4}$, batch size $128$, and a maximum of $100$ epochs with early stopping on the validation split. Hyperparameters are held constant across all models and all datasets. We do not tune per-dataset or per-model.

\paragraph{Transformer backbone.} The base ($L$-layer) transformer uses $L = 4$ layers, hidden dimension $256$, $8$ attention heads, FFN dimension $4 \times 256 = 1024$, and dropout $0.1$. The deeper baseline uses the same configuration with $2L = 8$ layers. Looped baselines reuse the same per-block configuration with $K \in \{1, 4\}$ and $M \in \{2, 4\}$.

\paragraph{CoT feedback MLP.} The MLP $\phi_\theta$ that maps query-position hidden states to feedback tokens is a two-layer MLP with hidden dimension $256$ and GELU activation, mapping from the transformer hidden dimension to itself.

\paragraph{Tabular setup.} We use a TabICL-style three-stage architecture~\cite{qu2025tabicl} with one column-attention layer, one row-attention layer, and a $4$-layer ICL transformer at the third stage. Recurrence is applied only at the ICL stage; the column and row stages are run once per query and their outputs cached across recurrences. Predictions are trained with cross-entropy for classification and root mean squared error for regression.

\paragraph{Time-series setup.} We use a patch-based forecasting transformer~\cite{nie2023patchtst} with patch size $16$, context length $1024$, and the prediction horizon set per dataset ($96$ for ETTh1/ETTh2/ETTm1/ETTm2/ECL/Traffic/Weather/Exchange, $24$ for ILI), following the standard LTSF-9 protocol~\cite{zeng2022lstf}. Forecasts are trained with the mean squared error.

\paragraph{Validation protocol.} For each dataset we hold out a validation split from the training data. The same split is used both for early stopping.

\section{nanoTabPFN-CoT pretraining details}
\label{app:nanotabpfn}

This appendix describes the architecture and pretraining setup used for the pretrained nanoTabPFN-CoT experiments reported in Section~\ref{sec:pretrained}. The setup differs from the from-scratch protocol in Appendix~\ref{app:training} in several respects, because the underlying architecture (TabPFN-v2 style, alternating row and feature attention) and pretraining regime (synthetic prior, large step count) are different.

\paragraph{Backbone.} We use the nanoTabPFN architecture~\cite{pfefferle2025nanotabpfn}, an open-source small-scale reimplementation of TabPFN-v2. The backbone is a 3-layer transformer with embedding dimension 96, 4 attention heads, and feedforward hidden dimension 192. Each block alternates feature attention (across the $d+1$ columns of a row, where $d$ is the number of features and the additional column carries the label) and datapoint attention (across rows). We use the default nanoTabPFN feature encoder (per-cell standardization using context-set statistics followed by a linear projection) and target encoder, and a 2-layer MLP decoder with GELU activation.

\paragraph{Adapting latent CoT to the TabPFN-v2 row layout.}
Unlike the from-scratch setup, where each CoT step appends one feedback token to a sequence, the TabPFN-v2 architecture represents each row as a sequence of $d+1$ vectors (one per feature, plus one for the label). A single CoT step therefore must append an entire row of $d+1$ vectors, not a single token. We append one feedback row per query: the compressed query-position hidden state occupies the label-column slot, and the $d$ feature-column slots are filled with a learned marker embedding shared across all CoT steps. Specifically, given the label-column hidden states $h_q \in \mathbb{R}^{B \times n_q \times E}$ at the query positions after a stack pass, the feedback row block $Z \in \mathbb{R}^{B \times n_q \times (d+1) \times E}$ is constructed by writing $\phi_\theta(h_q)$ into the label column and broadcasting the learned marker $m \in \mathbb{R}^E$ (initialized $\mathcal{N}(0, 0.02^2)$) into the $d$ feature columns. The feedback rows are appended along the row dimension and the full stack is re-run.

\paragraph{Group-aware datapoint attention.}
In the datapoint-attention sublayer, rows fall into three groups: context rows (with observed labels), query rows (the actual prediction targets), and feedback rows (added during CoT recurrence). To avoid leaking information between these groups in ways that break the in-context learning setup, attention contexts are configured as follows. Context rows attend only to other context rows. Query rows attend to context rows and feedback rows. Feedback rows attend to context, query, and other feedback rows. Feature attention is unchanged from nanoTabPFN and operates within each row independently. The first stack pass has no feedback rows and reduces to the standard TabPFN-v2 ICL setup.

\paragraph{Feedback MLP and gated residual.}
The feedback MLP $\phi_\theta$ is a 2-layer MLP (Linear $\to$ GELU $\to$ Linear) mapping from $E = 96$ to $E = 96$ with hidden dimension $96$. We initially observed stability issues when the recurrent passes were allowed to contribute to the residual stream at full strength from the start of training, so we introduced a single learned scalar gate $g = \tanh(g_0)$, with $g_0$ initialized to $2.0$, that modulates both the contents of the feedback rows and the overall CoT contribution. Concretely, the marker and the compressed query state in each feedback row are both scaled by $g$, and the final residual stream output combines the no-CoT pass with the recurrent passes as
\[
H_{\text{out}} = H^{(0)} + g \cdot (H^{(R)} - H^{(0)}),
\]
where $H^{(0)}$ is the stack output after the first pass and $H^{(R)}$ is the output after $R$ recurrent passes. The gate is differentiable but, in practice, $g$ remains close to 1 throughout training across all $R_{\text{train}}$ settings and seeds, so the learned model is effectively the un-gated CoT model; the parameterization is retained primarily for training stability.

\paragraph{Pretraining setup.}
We pretrain on the 300k-tasks TabICL prior dump used by \cite{pfefferle2025nanotabpfn}, available at \url{https://figshare.com/s/63fc1ada93e42e388e63}. Each model is trained for 4{,}000 steps with batch size 32 (one synthetic task per batch element). We use the schedule-free AdamW optimizer with learning rate $3 \times 10^{-4}$, no weight decay, and 250 warmup steps. Training uses bfloat16 mixed precision with gradient checkpointing across the recurrent stack passes to fit within GPU memory at higher $R$. Loss is cross-entropy loss. Gradients are clipped to a global norm of $1.0$.

\paragraph{Recurrence configuration.}
We pretrain three models with $R_{\text{train}} \in \{1, 2, 4\}$, each with 6 random seeds. Evaluation recurrence is matched to training recurrence (i.e., $R_{\text{eval}} = R_{\text{train}}$) in all pretrained-model experiments.

\paragraph{Evaluation.}
We evaluate on the binary-classification subset of TabArena used by the nanoTabPFN paper~\cite{pfefferle2025nanotabpfn}: OpenML tasks with at most 10 features, no missing values, at most 2 target classes, and a minority class fraction of at least 2.5\%. For each task we subsample at most 200 instances stratified by label to match the nanoTabPFN evaluation protocol, then evaluate with 5-fold stratified cross-validation and report mean ROC-AUC. We report mean and standard error across the 6 training seeds.

\section{Datasets}
\label{app:datasets}

\paragraph{Time series (LTSF-9).} We use the nine standard long-term forecasting datasets: ETTh1, ETTh2, ETTm1, ETTm2, ECL, Traffic, Weather, Exchange, and ILI, with horizons and splits following~\cite{zeng2022lstf}. Forecast horizon is $96$ for all datasets except ILI, which uses $24$.

\paragraph{Tabular (OpenML subset).} We use a 27-dataset subset of OpenML comprising 15 binary classification, 8 multiclass classification, and 4 regression datasets. The full list is given in Table \ref{tab:tabular_per_dataset}.

\section{Per-dataset results}
\label{app:per_dataset}

Tables \ref{tab:tabular_per_dataset} and \ref{tab:ts_per_dataset} report per-dataset performance for the same-depth baseline, deeper baseline, best CoT model (selected on validation across $R_\mathrm{train} \in \{1, 2, 4\}$ and $R_\mathrm{eval} \in \{0, 1, 2, 4, 8\}$), and best looped baseline (selected on validation across the four $K \times M$ configurations).

\begin{table*}[h]
\centering
\caption{Per-dataset results on LTSF-9 time-series forecasting. Metric is mean squared error of the median forecast (lower is better). Percent gain is relative to the same-depth baseline. CoT wins against the baseline on 7/9 datasets.}
\label{tab:ts_per_dataset}
\resizebox{\linewidth}{!}{
\begin{tabular}{lrrrrrrr}
\toprule
Dataset & Horizon & Baseline & Deeper & CoT $(R=2)$ & Best Looped $(K=4)$ & CoT \% gain & Looped \% gain \\
\midrule
ETTh1     & 96 & $0.5166 \pm 0.0520$ & $0.5773 \pm 0.0282$ & $0.4176 \pm 0.0177$ & $0.5014 \pm 0.0019$ & $+23.99\% \pm 11.70\%$ & $+8.47\% \pm 14.99\%$ \\
ETTh2     & 96 & $0.4011 \pm 0.0687$ & $0.5632 \pm 0.1107$ & $0.2788 \pm 0.0300$ & $0.3558 \pm 0.0700$ & $+28.97\% \pm 4.49\%$ & $+10.99\% \pm 13.21\%$ \\
ETTm1     & 96 & $0.5933 \pm 0.0210$ & $0.6289 \pm 0.0519$ & $0.5685 \pm 0.0547$ & $0.5905 \pm 0.0203$ & $+4.56\% \pm 6.15\%$ & $+0.27\% \pm 4.36\%$ \\
ETTm2     & 96 & $0.3656 \pm 0.0560$ & $0.3422 \pm 0.0105$ & $0.2767 \pm 0.0574$ & $0.3741 \pm 0.0518$ & $+22.80\% \pm 12.88\%$ & $-3.59\% \pm 7.24\%$ \\
ECL       & 96 & $0.3198 \pm 0.0083$ & $0.3208 \pm 0.0132$ & $0.3684 \pm 0.0088$ & $0.3259 \pm 0.01318$ & $-15.21\% \pm 0.54\%$ & $-2.15\% \pm 5.77\%$ \\
Traffic   & 96 & $0.551 \pm 0.006$ & $0.538 \pm 0.002$ & $0.544 \pm 0.003$ & $0.538 \pm 0.006$ & $-1.41\% \pm 1.15\%$ & +$2.22\% \pm 1.47\%$ \\
Weather   & 96 & $0.1748 \pm 0.0047$ & $0.1689 \pm 0.0036$ & $0.1684 \pm 0.0014$ & $0.1685 \pm 0.0020$ & $+3.56\% \pm 1.85\%$ & $+3.53\% \pm 1.53\%$ \\
Exchange  & 96 & $1.1844 \pm 0.0356$ & $1.1299 \pm 0.0791$ & $0.8910 \pm 0.0191$ & $1.1605 \pm 0.0921$ & $+24.63\% \pm 2.95\%$ & $+1.70\% \pm 9.17\%$ \\
ILI       & 24 & $2.20 \pm 0.029$ & $2.36 \pm 0.02$ & $1.68 \pm 0.048$ & $2.28\ \pm 0.017$ & $+21.79\% \pm 4.13\%$ & $+3.48\% \pm 1.92\%$ \\
\midrule
\textbf{Average} & & & & & & $\mathbf{+12.63\% \pm 5.08\%}$ & $\mathbf{+2.77\% \pm 1.05\%}$ \\
\bottomrule
\end{tabular}
}
\end{table*}

\begin{table*}[h]
\centering
\caption{Per-dataset results on 27 datasets from OpenML \cite{openml}. Metric is AUC for binary classification, accuracy for multiclass classification, and $-$RMSE for regression (higher is better in all cases). Percent gain is relative to the same-depth baseline. CoT wins against the baseline on 23/27 datasets.}
\label{tab:tabular_per_dataset}
\resizebox{\linewidth}{!}{
\begin{tabular}{llrrrrrr}
\toprule
Dataset & Task & Baseline & Deeper & CoT $(R=1)$ & Looped $(K=4, M=2)$ & CoT \% gain & Looped \% gain \\
\midrule
balance\_scale       & multiclass & $0.9471 \pm 0.019$ & $0.963 \pm 0.023$ & $0.9945 \pm 0.029$ & $0.9312 \pm 0.023$ & $+5.00\% \pm 3.76\%$ & $-1.68\% \pm 3.14\%$ \\
breast\_w            & binclass   & $0.9912 \pm 0.0066$ & $0.99 \pm 0.007$ & $0.9942 \pm 0.0064$ & $0.9816 \pm 0.0075$ & $+0.3\% \pm 0.45\%$ & $-0.97\% \pm 1.00\%$ \\
cmc                  & multiclass & $0.5563 \pm 0.006$ & $0.5676 \pm 0$ & $0.5723 \pm 0.0047$ & $0.5743 \pm 0.022$ & $+2.87\% \pm 1.40\%$ & $+3.24\% \pm 4.06\%$ \\
credit\_g            & binclass   & $0.7335 \pm 0.0091$ & $0.721 \pm 0.013$ & $0.7619 \pm 0.021$ & $0.7444 \pm 0.0086$ & $+3.87\% \pm 3.19\%$ & $+1.49\% \pm 1.72\%$ \\
diabetes             & binclass   & $0.857 \pm 0.015$ & $0.841 \pm 0.017$ & $0.8958 \pm 0.018$ & $0.8519 \pm 0.011$ & $+4.52\% \pm 2.83\%$ & $-0.60\% \pm 2.18\%$ \\
tic\_tac\_toe        & binclass   & $0.6527 \pm 0.041$ & $0.6859 \pm 0.029$ & $0.7065 \pm 0.032$ & $0.6051 \pm 0.037$ & $+8.25\% \pm 8.36\%$ & $-7.29\% \pm 8.09\%$ \\
vehicle              & multiclass & $0.6784 \pm 0.0039$ & $0.6235 \pm 0.0068$ & $0.7 \pm 0.044$ & $0.6196 \pm 0.021$ & $+3.19\% \pm 6.45\%$ & $-8.67\% \pm 3.11\%$ \\
eucalyptus           & multiclass & $0.491 \pm 0.012$ & $0.5135 \pm 0.028$ & $0.5014 \pm 0.0095$ & $0.4775 \pm 0.012$ & $+2.11\% \pm 3.14\%$ & $-2.75\% \pm 3.38\%$ \\
wdbc                 & binclass   & $0.9665 \pm 0.0069$ & $0.959 \pm 0.0085$ & $0.9962 \pm 0.012$ & $0.9683 \pm 0.0088$ & $+3.87\% \pm 1.45\%$ & $+0.19\% \pm 1.16\%$ \\
banknote             & binclass   & $0.8609 \pm 0.02$ & $0.8459 \pm 0.028$ & $0.8858 \pm 0.029$ & $0.8565 \pm 0.009$ & $+2.89\% \pm 4.18\%$ & $-0.51\% \pm 2.57\%$ \\
blood\_transfusion   & binclass   & $0.7411 \pm 0.014$ & $0.7502 \pm 0.015$ & $0.7679 \pm 0.015$ & $0.7541 \pm 0.022$ & $+3.61\% \pm 2.84\%$ & $+1.75\% \pm 3.58\%$ \\
kr\_vs\_kp           & binclass   & $0.9745 \pm 0.0075$ & $0.976 \pm 0.0083$ & $0.9912 \pm 0.016$ & $0.9812 \pm 0.0073$ & $+1.6\% \pm 0.86\%$ & $+0.69\% \pm 1.08\%$ \\
phoneme              & binclass   & $0.8678 \pm 0.0065$ & $0.8648 \pm 0.0092$ & $0.9136 \pm 0.0085$ & $0.8808 \pm 0.0033$ & $+5.28\% \pm 1.26\%$ & $+1.50\% \pm 0.85\%$ \\
qsar\_biodeg         & binclass   & $0.8716 \pm 0.012$ & $0.8843 \pm 0.0074$ & $0.9029 \pm 0.018$ & $0.8708 \pm 0.0088$ & $+3.59\% \pm 2.46\%$ & $-0.09\% \pm 1.68\%$ \\
wall\_robot          & multiclass & $0.7656 \pm 0.03$ & $0.746 \pm 0.014$ & $0.8205 \pm 0.018$ & $0.7173 \pm 0.017$ & $+7.17\% \pm 4.86\%$ & $-6.31\% \pm 4.33\%$ \\
phishing             & binclass   & $0.8569 \pm 0.031$ & $0.8437 \pm 0.039$ & $0.8774 \pm 0.036$ & $0.8081 \pm 0.0092$ & $+2.39\% \pm 5.67\%$ & $-5.69\% \pm 3.62\%$ \\
steel\_plates        & multiclass & $0.6632 \pm 0.02$ & $0.6684 \pm 0.012$ & $0.6748 \pm 0.0095$ & $0.6547 \pm 0.036$ & $+1.75\% \pm 3.34\%$ & $-1.28\% \pm 6.14\%$ \\
segment              & multiclass & $0.6681 \pm 0.049$ & $0.7258 \pm 0.016$ & $0.7152 \pm 0.024$ & $0.7042 \pm 0.0038$ & $+7.04\% \pm 8.67\%$ & $+5.40\% \pm 7.77\%$ \\
churn                & binclass   & $0.6893 \pm 0.052$ & $0.7102 \pm 0.027$ & $0.7178 \pm 0.054$ & $0.728 \pm 0.0056$ & $+4.13\% \pm 11.09\%$ & $+5.61\% \pm 7.99\%$ \\
electricity          & binclass   & $0.7898 \pm 0.0042$ & $0.7883 \pm 0.0069$ & $0.8303 \pm 0.0065$ & $0.7975 \pm 0.0078$ & $+5.13\% \pm 1.00\%$ & $+0.97\% \pm 1.12\%$ \\
adult                & binclass   & $0.8179 \pm 0.004$ & $0.8288 \pm 0.013$ & $0.852 \pm 0.016$ & $0.8173 \pm 0.012$ & $+4.17\% \pm 2.04\%$ & $-0.07\% \pm 1.50\%$ \\
bank\_marketing      & binclass   & $0.8542 \pm 0.0005$ & $0.8483 \pm 0.0084$ & $0.8921 \pm 0.011$ & $0.8587 \pm 0.0047$ & $+4.43\% \pm 1.30\%$ & $+0.53\% \pm 0.55\%$ \\
jungle\_chess        & multiclass & $0.7241 \pm 0.004$ & $0.72 \pm 0.017$ & $0.7735 \pm 0.0039$ & $0.7367 \pm 0$ & $+6.83\% \pm 0.80\%$ & $+1.74\% \pm 0.56\%$ \\
california\_housing  & regression & $-0.2062 \pm 0.0013$ & $-0.2032 \pm 0.0017$ & $-0.2121 \pm 0.005$ & $-0.2049 \pm 0.0048$ & $-2.86\% \pm 2.53\%$ & $+0.63\% \pm 2.41\%$ \\
abalone              & regression & $-2.487 \pm 0.012$ & $-2.444 \pm 0.0008$ & $-2.51 \pm 0.038$ & $-2.323 \pm 0.041$ & $-0.94\% \pm 1.59\%$ & $+6.60\% \pm 1.70\%$ \\
cpu\_act             & regression & $-152.2 \pm 0.22$ & $-153.6 \pm 1.3$ & $-160.8 \pm 0.5$ & $-152.2 \pm 0.78$ & $-5.64\% \pm 0.37\%$ & $+0.04\% \pm 0.53\%$ \\
superconductivity    & regression & $-13.24 \pm 0.96$ & $-12.96 \pm 0.62$ & $-13.91 \pm 0.75$ & $-13.88 \pm 1$ & $-5.10\% \pm 9.51\%$ & $-4.88\% \pm 10.85\%$ \\
\midrule
\textbf{Average} & & & & & & $\mathbf{+3.25\% \pm 0.88\%}$ & $\mathbf{-0.39\% \pm 0.80\%}$ \\
\bottomrule
\end{tabular}
}
\end{table*}

\end{document}